\documentclass[journal]{IEEEtran}
\ifCLASSINFOpdf
\else
\fi
\usepackage{url}
\usepackage{times}
\usepackage{soul}
\usepackage{color}
\usepackage[colorlinks,
            linkcolor=red,
            anchorcolor=blue, citecolor=green]{hyperref}
\usepackage[utf8]{inputenc}
\usepackage{graphicx}

\usepackage{amsmath}
\usepackage{amsthm}
\usepackage{booktabs}
\usepackage{algorithm}
\usepackage{algorithmic}

\usepackage{colortbl}

\usepackage{amssymb}
\usepackage{multirow}
\usepackage{multicol}
\usepackage{comment}
\definecolor{mygray}{gray}{.97}

\begin{document}
\title{Beyond Instance Discrimination: Relation-aware Contrastive Self-supervised Learning}

\author{Yifei~Zhang,
        Chang~Liu,
        Yu~Zhou,
        Weiping~Wang,
        Qixiang~Ye,
        and~Xiangyang~Ji% <-this % stops a space
% \thanks{}
\thanks{Y. Zhang, Y. Zhou and W. Wang are with the Institute of Information Engineering, Chinese Academy of Sciences, also with the School of Cyber Security, University of Chinese Academy of Sciences, Beijing 100089, China, E-mail: \{zhangyifei0115, zhouyu, wangweiping\}@iie.ac.cn. Y. Zhou is the corresponding author. Q. Ye is with the School of Electronic, Electrical, and Communication Engineering, University of Chinese Academy of Sciences, Beijing 101408, China, E-mail: qxye@ucas.ac.cn. C. Liu and X. Ji are with the Department of Automation, Tsinghua University, Beijing 100084, China, E-mail: \{liuchang2022, xyji\}@tsinghua.edu.cn. Y. Zhang and C. Liu contribute equally.}% <-this % stops a space
%\thanks{Q. Ye is with the School of Electronic, Electrical and Communication Engineering, University of Chinese Academy of Sciences, Beijing 101408, China, (e-mail: qxye@ucas.ac.cn).}% <-this % stops a space
%\thanks{W. Wang is with Institute of Information Engineering, Chinese Academy of Sciences, Beijing 100089, China, E-mail: wangweiping@iie.ac.cn.}
%\thanks{X. Ji is with the Department of Automation, Tsinghua University, Beijing 100084, China, (e-mail: xyji@tsinghua.edu.cn).}% <-this % stops a space
%\thanks{W. Wang was with the Institute of Information Engineering, Chinese Academy of Sciences, Beijing 100093, China, e-mail:  wangweiping@iie.ac.cn.}% <-this % stops a space
%\thanks{Y. Zhang and C. Liu contribute equally.}
%\thanks{Y. Zhou is the corresponding author.}
        }

\maketitle

\begin{abstract}
Contrastive self-supervised learning (CSL) based on instance discrimination typically attracts positive samples while repelling negatives to learn representations with pre-defined binary self-supervision.
However, vanilla CSL is inadequate in modeling sophisticated instance relations, limiting the learned model to retain fine semantic structure.
On the one hand, samples with the same semantic category are inevitably pushed away as negatives. On the other hand, differences among samples cannot be captured.
In this paper, we present relation-aware contrastive self-supervised learning (ReCo) to integrate instance relations, \textit{i.e.}, global distribution relation and local interpolation relation, into the CSL framework in a plug-and-play fashion.
Specifically, we align similarity distributions calculated between the positive anchor views and the negatives at the global level to exploit diverse similarity relations among instances.
Local-level interpolation consistency between the pixel space and the feature space is applied to quantitatively model the feature differences of samples with distinct apparent similarities. 
Through explicitly instance relation modeling, our ReCo avoids irrationally pushing away semantically identical samples and carves a well-structured feature space.
Extensive experiments conducted on commonly used benchmarks justify that our ReCo consistently gains remarkable performance improvements.

\end{abstract}

\begin{IEEEkeywords}
Global distribution relation, local interpolation relation, relation-aware contrastive self-supervised learning, self-supervised learning.
\end{IEEEkeywords}

% \fi
\IEEEpeerreviewmaketitle

\section{Introduction}
\IEEEPARstart{I}{n} the deep learning era, large-scale pre-training~\cite{ImageNet-IJCV15,Instagram} then downstream fine-tuning has become a dominant learning paradigm~\cite{alexnet,ResNet,TMM18_representation01,TMM21_representation02}.
However, supervised pre-training typically focuses on task-specific features, resulting in limited model generalization. Building finely annotated large-scale datasets is also laborious, expensive, and sometimes impractical. 
Inspired by human cognition from unlabeled data, unsupervised visual representation learning is attracting growing attention~\cite{Dosovitskiy_2016_Exemplar,Zhang_2019_AET,Qi_2019_AVT,He_2020_Moco,hcl_tmm21,ldz1,ldz2,lxn2,liu2022self,surveySSL_tpami,QGJ_pami_smalldata,QJG_pami_AET}.

Mainstream approaches either manually design specific pretext tasks
to assimilate the intrinsic data structure~\cite{Doersch_2015_Context, Noroozi_2016_Jigsaw, Guatav_2016_Color, Inpainting_2016, Gidaris_2018_Rotnet}, or encode data similarities with a contrastive self-supervised learning (CSL) paradigm~\cite{Oord_2018_CPC,Wu_2018_IR,He_2020_Moco,Chen_2020_SimCLR}.
Unlike handcrafted pretext tasks that are limited in exhausting correlating human priori, CSL with instance discrimination aims at learning view-invariant representation, which presents superior performance and great potential~\cite{Ye_2019_IS,He_2020_Moco,Chen_2020_SimCLR,analysis_cvpr21,understandeCL_CVPR21}.
Based on InfoNCE loss~\cite{Oord_2018_CPC}, ISIF~\cite{Ye_2019_IS}, MoCo~\cite{He_2020_Moco} and SimCLR~\cite{Chen_2020_SimCLR} introduce siamese networks to attract different instance views as positives while repelling other instances in a mini-batch or a memory bank as negatives.
However, since negative samples are naively defined as different images, false negatives with the same semantic content inevitably occur, and their specific similarity relations are also not taken into account.
Models learned with ``hard" binary positive and negative assignments are apparently limited by biased and incomplete semantic structure learning of the data.

%%%%%%%%%%%%%%%%%%%%%%%%%%
\begin{figure}[!t]
\centering
% width=1.0\columnwidth
\includegraphics[width=\columnwidth]{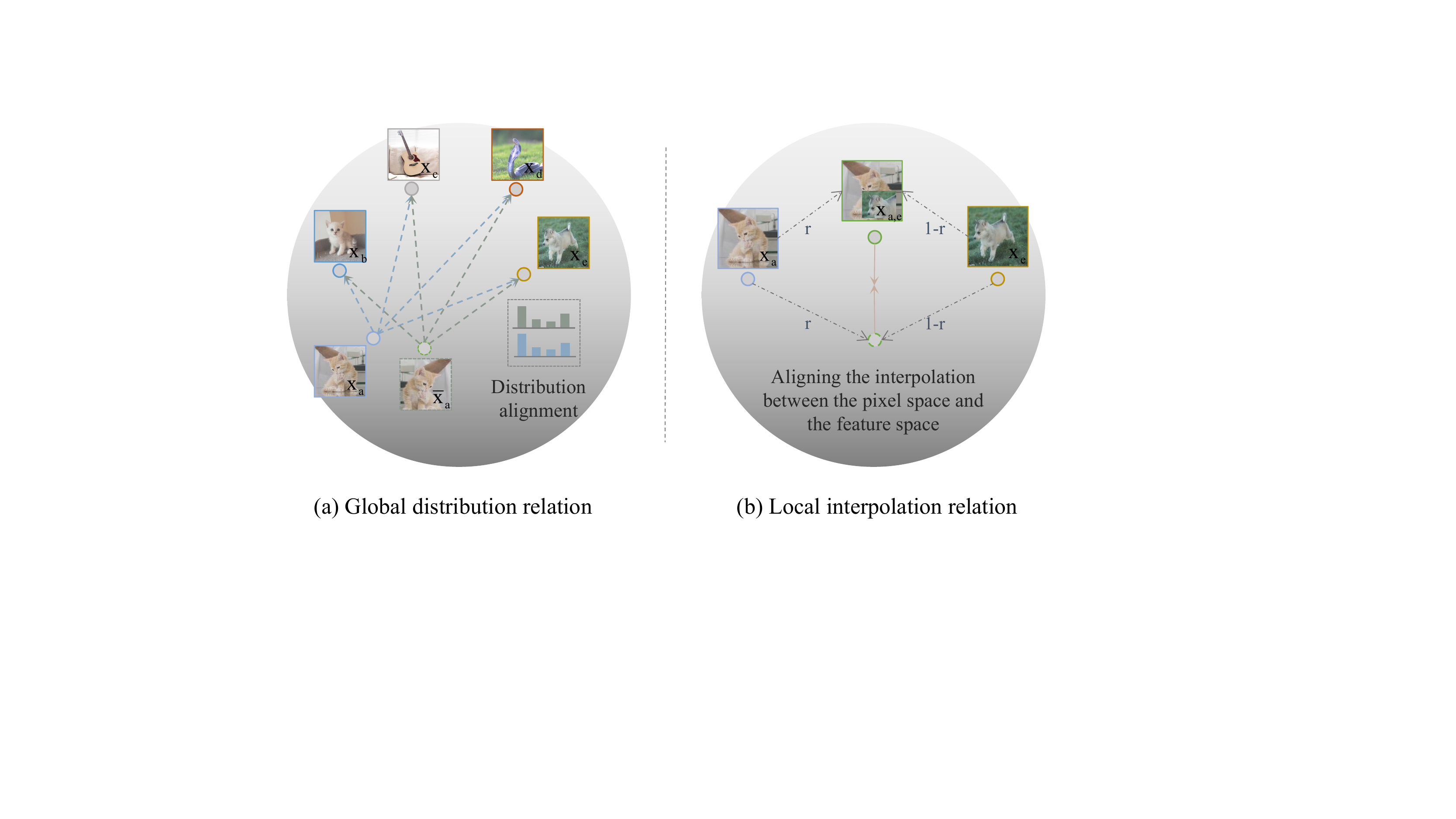}

\caption[]{Instance relation illustration:
(a) Global distribution relation enriches the view-invariant representation from the instance pair level to the dataset level by aligning similarity distributions. 
Specific similarities for negative pairs are well-exploited, for example, the similarity between cat $x_a$ and cat $x_b$ is higher than cat $x_a$ and guitar $x_c$. 
(b) Local interpolation relation quantitatively controls the apparent similarity by utilizing a data mixture technique and exploits the interpolation consistency by aligning the interpolation between the pixel space and the feature space. 
The color-circled points in the figure indicate the features corresponding to the images.}
\label{fig:fig1}
\end{figure}

\begin{figure*}
\centering
% width=1.0\columnwidth
\includegraphics[width=\linewidth]{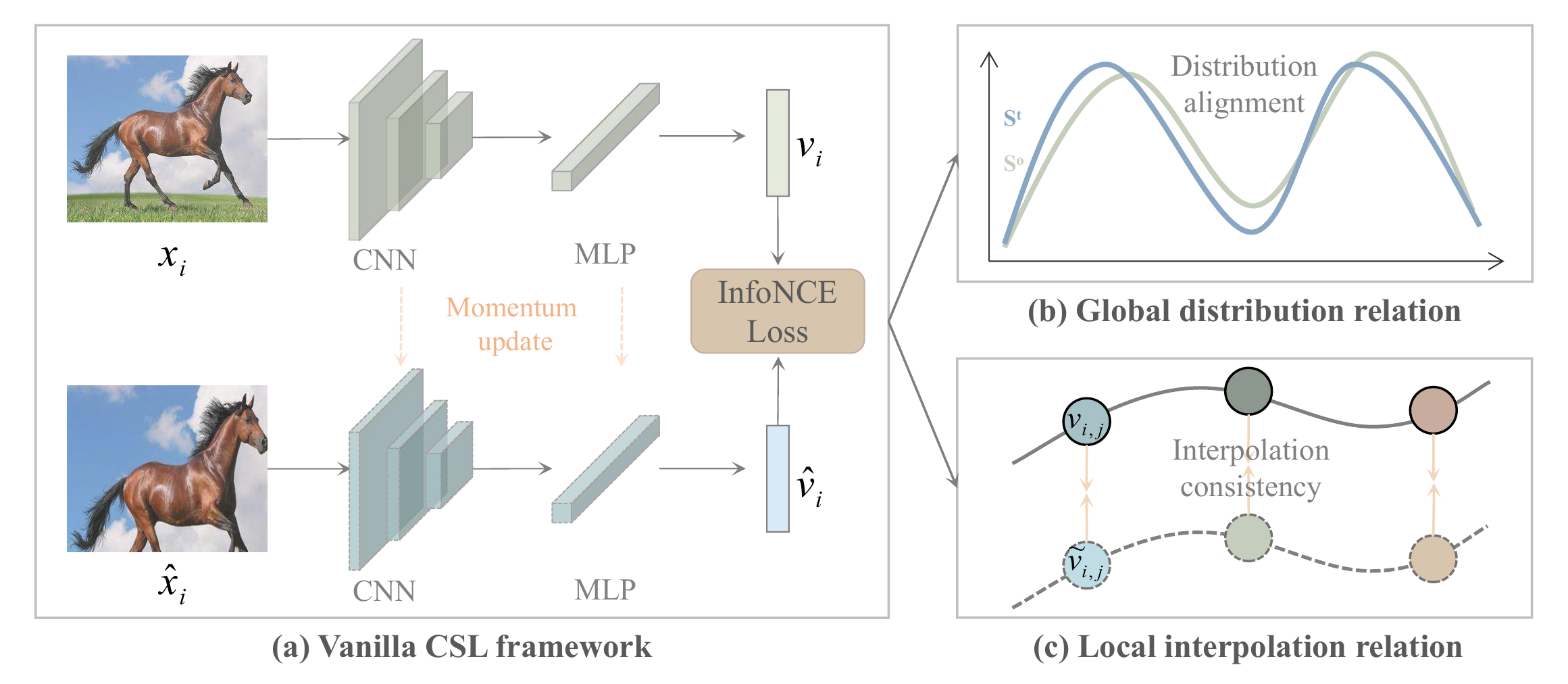}
\caption[]{Architecture overview of our ReCo, which consists of (a) a vanilla CSL framework, \textit{e.g.}, MoCo-v2, (b) global distribution relation, and (c) local interpolation relation. $v_i$, $\hat{v}_i$, and $v_{i,j}$ are features of input views $x_i$ and $\hat{x}_i$, and the interpolated image, respectively. $\Tilde{v}_{i,j}$ denotes the interpolated feature.}

\label{fig:fig2}
\end{figure*}

In this paper, we propose a simple yet effective \textbf{re}lation-aware \textbf{co}ntrastive self-supervised learning (ReCo) approach to concurrently explore ``soft" instance relations of global distribution and local interpolation, Figure~\ref{fig:fig1}.
Specifically, in the global perspective, we enrich positive sample pairs with positive distribution pairs by calculating similarity distributions of augmented input views to their negative samples. Feature representation can be significantly improved by explicitly coupling complex similarity information between the positive augmented samples and the negative samples with distribution alignment,
Figure~\ref{fig:fig2}(b).
In the local perspective, we interpolate randomly selected images in a mini-batch with a typical data mixture strategy, \textit{e.g.}, cutmix~\cite{cutmix_2019_iccv}.
The interpolation ratio can quantitatively control the apparent similarity of the synthetic image to the original image pair.
Meanwhile, we interpolate features of the image pair
with the same ratio to obtain the feature as the self-supervision signal of the interpolated image. Attracting corresponding features in the feature space, the consistency of local interpolation relation can be assimilated, Figure~\ref{fig:fig2}(c).

By incorporating the global distribution and local interpolation relations in a plug-and-play fashion, the proposed ReCo takes full use of specific similarities of diverse sample pairs to relax the constraint that all positives/negatives should be equally attracted/repelled.
Extensive experiments justify the effectiveness of ReCo, which produces a locally aggregated yet globally uniform feature space, Figure~\ref{fig:fig5}.
Specifically, ReCo achieves state-of-the-art performance with 75.9\% top-1 accuracy for linear classification and 78.9\% and 87.9\% top-5 accuracies for semi-supervised classification with 1\% and 10\% labeled data.
Transferring to the VOC~\cite{VOC} dataset, ReCo improves MoCo-v2~\cite{mocov2_arXiv2020} by at least 6.3\% mAP for low-shot classification with k=1,2,4,8,16 and 0.9\% AP$_{50}$ for object detection.

The contributions are summarized as follows: 
\begin{enumerate}
        \item We propose relation-aware contrastive self-supervised learning (ReCo) to effectively retain the data semantic structures by exploring instance relations from both global and local perspectives. It is a novel attempt to break through the limitation of the error-prone binary label assignment of vanilla CSL.
		
		\item We exploit the global distribution relation to explicitly constrain the specific similarity of different samples other than repelling all negative samples equally. 
		
		\item We exploit the local interpolation relation to carve the semantic structure of the feature space with quantitative appearance similarity retention.

		\item The proposed ReCo outperforms existing CSL works on multiple benchmarks and shows better generalization ability, especially for insufficient supervision regimes, \textit{e.g.}, it significantly exceeds MoCo-v2 in semi-supervised learning with 1\%/10\% labeled data and low-shot classification with 1/2/4/8/16 samples.
		
\end{enumerate}

%%%%%%%%%%%%%%%%%%%%%%%%%%%%%%%%%%%%%%%%%%%%%

\section{Related Works}

\subsection{Unsupervised Visual Representation Learning}
Unsupervised visual representation learning aims at utilizing unlabeled data to learn transferable feature representations to initialize downstream tasks, such as image classification~\cite{ResNet}, object detection~\cite{fasterrcnn,LTM_tpami}, and semantic segmentation~\cite{Cityscapes,TMM22_segmentation}, which can be roughly divided into handcrafted pretext tasks and contrastive self-supervised learning.

{\bf Handcrafted Pretext Tasks.}
Such methods typically assimilate common sense through self-supervision signals generated based on the inherent structure of the data.
Specifically, some works aim at recovering input images under pre-defined corruptions, such as colorization~\cite{Guatav_2016_Color}, inpainting~\cite{Inpainting_2016}, and split-brain autoencoding~\cite{Zhang_2017_splitbrain}.  
Some works generate self-supervision via specific transformations, such as context prediction~\cite{Doersch_2015_Context}, solving jigsaw puzzle~\cite{Noroozi_2016_Jigsaw}, rotation prediction~\cite{Gidaris_2018_Rotnet}, \textit{etc}. 
Developing sophisticated pretext tasks largely depends on human prior knowledge, which limits their rapid evolution.

%%%
{\bf Contrastive Self-supervised Learning.}
With InfoNCE loss~\cite{Oord_2018_CPC} and its variants, CSL methods typically construct informative positive and negative sets to encode similarities of positive instance pairs and differences of negative ones.
NPID~\cite{Wu_2018_IR} introduces a memory bank to store features of the whole dataset and formulates instance discrimination~\cite{Dosovitskiy_2016_Exemplar} as a non-parametric classification problem.
MoCo~\cite{He_2020_Moco} proposes a moving-average encoder and a dynamic queue to build positive and negative pairs effectively and efficiently. SimCLR~\cite{Chen_2020_SimCLR} fulfills the contrast procedure in the current mini-batch and introduces more data augmentations to report impressive performance. 
Interestingly, some researches justify that augmentation invariant representations can also be well learned without negative samples, such as BYOL~\cite{BYOL_nips20}, SimSiam~\cite{SimSiam_CVPR21}, SwAV~\cite{SwAv_nips20}, BarlowTwins~\cite{BarlowTwins_arXiv2021}, \textit{etc}. Moreover, some 
works~\cite{Feng_2019_Decoupling,jigclu_cvpr21} attempt to combine contrastive loss with handcrafted pretext tasks, which demonstrate their complementary nature.

%%%%%%%%%%%%%%%%%%%%%%%%%%%%%%%%%%%%%%%%%%%%%
To better explore class boundary information, some recent works delve into positive sample discovery. Clustering-based methods~\cite{Yang_2016_CVPR, Caron_2018_DC, Zhan_2020_ODC, ICLR20_SeLa, PCL_iclr21,cluster_tmm19} target at iteratively grouping instances for reliable pseudo label assignment. Neighbour-discovery-based methods~\cite{Huang_2019_AND, PRrelation-MilbichGDO20, DPSIS_TCYB21, InvPro_nips20, ICCV21_NNCLR} usually set specific rules to select reliable positive samples in the local neighbourhood. 
However, as a strong addition to CSL, relation-aware contrastive learning based on soft instance relations of similarity distribution at the global level and interpolation consistency at the local level has not been fully exploited, which hinders the development of CSL.

\subsection{Instance Relations Exploration}

The informative data semantic structure can be captured via instance relation exploration, which is usually established in terms of similarity distribution and data interpolation~\cite{hinton_distillation,mixup_2018_iclr,ICLR21_CO2,ReSSL_nips21,Unmix_AAAI22}. The distribution depicts unique similarities of diverse sample pairs and the interpolation consistency models relations between synthetic images and original inputs. Their complementary nature appears under-studied.

\textbf{Similarity Distribution.}
Similarity distribution is typically exploited in knowledge distillation~\cite{hinton_distillation} and consistency regularization in semi-supervised learning~\cite{temporalEnsembling_iclr17,meanteacher_nips17}.
Logit-based knowledge distillation~\cite{hinton_distillation} proposes to use the output of the softmax layer of the teacher model as soft labels to train the student model.
Its effectiveness lies in the fact that the soft labels depict the relation between different classes.
After that, some methods explicitly establish the structural relation between the outputs of different samples rather than individual outputs themselves, \textit{e.g.},
relational knowledge distillation~\cite{cvpr19_RKD}, similarity-preserving knowledge distillation~\cite{iccv19_SP}, and self-supervised distillation~\cite{SEED_ICLR2021}, \textit{etc.}

Consistency regularization in semi-supervised learning~\cite{meanteacher_nips17,mixmatch_nips19} insists that the output of the model should be similar before and after perturbing the input data, which is achieved by distribution alignment. 
A lot of semi-supervised learning works are devoted to how to generate better target distribution, \textit{e.g.}, Mean Teacher~\cite{meanteacher_nips17}, MixMatch~\cite{mixmatch_nips19}, SsCL~\cite{TMM22_semiContrastive}, \textit{etc}.
Some current self-supervised learning methods~\cite{ICLR21_CO2,ICCV21_ISD,ReSSL_nips21,clsa_arxiv21} are exploring the utilization of similarity distribution and have achieved remarkable results.
Typically, CO2~\cite{ICLR21_CO2} improves MoCo-v2 by additionally aligning the similarity distribution of two views to negative samples.
ISD~\cite{ICCV21_ISD} and ReSSL~\cite{ReSSL_nips21} utilize weak data augmentation to optimize the distribution alignment term without explicitly pushing away negative samples. 
CLSA~\cite{clsa_arxiv21} matches the distribution obtained from stronger and regular augmentations to explore new patterns ignored in MoCo-v2.
However, the local-level relation that apparently similar inputs should be close in feature space is not explicitly considered.

\textbf{Data Mixture.}
Data mixture typically targets at augmenting the sample space to reduce incompatibilities during inference. The model generalization ability can be enhanced by exploring relations between synthetic and raw data. 
Mixup~\cite{mixup_2018_iclr} performs the corresponding pixel-weighted summation of the input image pairs, and the label is also linearly interpolated. 
CutMix~\cite{cutmix_2019_iccv} replaces the removed regions with a patch from another image. 
Beyond supervised scenarios, data mixture is also applied in semi-supervised~\cite{ijcai19_ICT} and unsupervised~\cite{Unmix_AAAI22} learning. 
Specifically, UnMix~\cite{Unmix_AAAI22} and MixCo~\cite{arxiv20_mixco} perform data mixing in the input space, and then weight the loss with the interpolation ratio. Deviating from merely local-level interpolation, ReCo further exploits the similarity distribution to delineate global-level relations.

\section{Methodology}
\label{sec:method}

CSL methods based on instance discrimination typically rely on predefined hard binary assignments, which are error-prone and ignore the exploitation of different relations among instances. To retain the semantic structure of the data and produce a locally aggregated and globally uniform feature space~\cite{icml20_alignment_uniform,understandeCL_CVPR21}, we propose relation-aware contrastive self-supervised learning (ReCo) which simultaneously explores soft instance relations of similarity distribution at the global level and interpolation consistency at the local level, Figure~\ref{fig:fig2}.

\subsection{Overview}

% Introducing MoCo
\textbf{Baseline.} 
We choose the seminal work MoCo-v2~\cite{mocov2_arXiv2020} to clarify the implementation details of our ReCo which can also be applied on common CSL frameworks. It takes two views of the $i$-th instance $x_{i}$ and $\hat{x}_{i}$ as input, which are generated from the same image through a combination of data augmentations. The corresponding features $v_{i}$ and $\hat{v}_{i}$ are extracted by an online encoder $f_{\theta}$ and a momentum encoder $f_{\hat{\theta}}$ as $v_{i}=f_{\theta}(x_{i})$ and $\hat{v}_{i}=f_{\hat{\theta}}(\hat{x}_{i})$, where the encoder consists of a backbone network ($e.g.$ ResNet-50~\cite{ResNet}) and an MLP head, Figure~\ref{fig:fig2}(a). The feature $\hat{v}_{i}$ from the momentum branch is stored in a queue (memory bank) with the size of $K$. $v_{i}$ and $\hat{v}_{i}$ are defined as positive sample pairs that attract each other in the feature space while staying away from the negative samples in the queue. The learning objective is to minimize the InfoNCE~\cite{Oord_2018_CPC} loss:
\begin{equation}
    \mathcal{L}_{csl} = - \frac{1}{N} \sum_{i=1}^{N} log \frac{exp(v_{i}\cdot \hat{v}_{i} / \tau)}{exp(v_{i}\cdot \hat{v}_{i} / \tau) + \sum_{j}^{K} exp(v_{i} \cdot \Tilde{v}_{j} / \tau)},
    \label{mocoloss}
\end{equation}
where $N$ is the training set size, $\tau$ is the temperature~\cite{hinton_distillation}, and $\Tilde{v}_{j}$ is the $j$-th sample in the queue.

%%%%%%%%%%%%%%%%%%%%%%%%%%
\begin{figure}[!t]
\centering
% width=1.0\columnwidth
\includegraphics[width=\columnwidth]{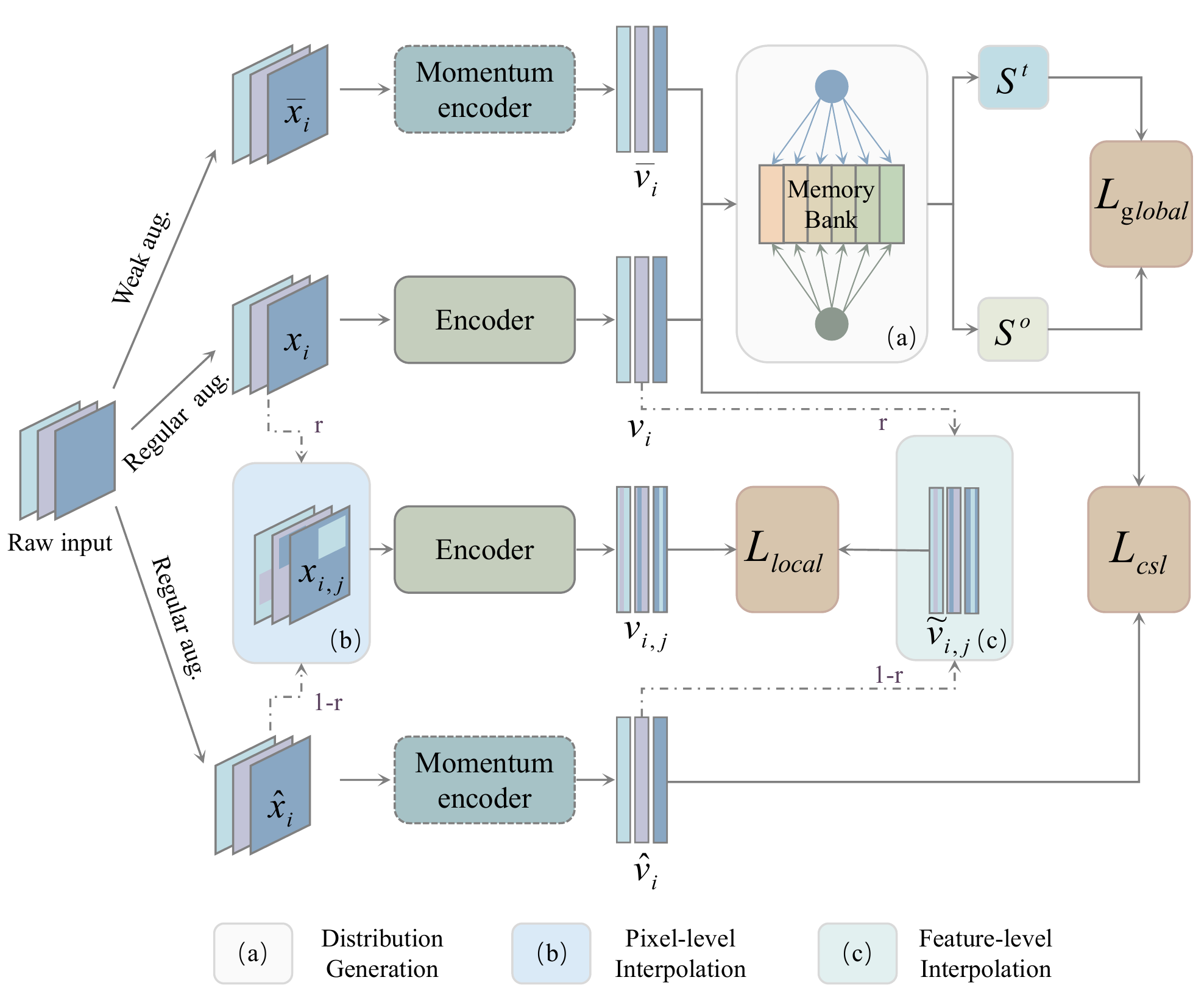}
\caption[]{Detailed implementation framework of our ReCo, in which three modules are jointly optimized.
}
\label{fig:fig3}
\end{figure}

{\bf Pipeline.} 
As illustrated in Figure~\ref{fig:fig2}, ReCo consists of three modules : (a) a vanilla CSL framework (MoCo-v2), (b) global distribution relation, and (c) local interpolation relation.
The global distribution relation utilizes distribution alignment to fully use the specific similarities of diverse samples to relax the constraint that all negatives are equally repelled.
The local interpolation relation applies image interpolation between random sample pairs. It explores the interpolation consistency relation between pixel and feature space to quantitatively model samples' apparent similarity.

The overall loss function of ReCo is a combination of the infoNCE loss $\mathcal{L}_{csl}$, the global distribution relation loss $\mathcal{L}_{global}$, and the local interpolation relation loss $\mathcal{L}_{local}$, which can be formulated as
% \begin{equation}
%      \mathop{\arg\min}\limits_{\theta} \mathcal{L} =  \mathcal{L}_{csl} + \lambda_{1}  \mathcal{L}_{global} + \lambda_{2}  \mathcal{L}_{local},
%     \label{l-all}
% \end{equation}
% \textcolor{blue}{no argmin}
\begin{equation}
      \mathcal{L} =  \mathcal{L}_{csl} + \lambda_{1}  \mathcal{L}_{global} + \lambda_{2}  \mathcal{L}_{local},
    \label{l-all}
\end{equation}
where $\lambda_{1}$ and $\lambda_{2}$ are balance weights of $\mathcal{L}_{global}$ and $\mathcal{L}_{local}$.

\subsection{Global Distribution Relation}
We extend view-invariant representation learning from the instance level to the distribution level, which is inspired by the consistency regularization in semi-supervised learning that the output of the model (probability distribution) should be similar under variations in the input space~\cite{meanteacher_nips17,mixmatch_nips19}. The distribution depicts the specific similarities between different classes and therefore retains rich global relations. Concretely, the global distribution relation is materialized with distribution generation and distribution alignment, Figure~\ref{fig:fig3}. 

\textbf{Distribution Generation.}
We calculate the similarity distribution of each input view to its negative samples based on the embedding features extracted by the encoder.
To obtain a stable target distribution, we employ weak data augmentation that does not introduce severe variation~\cite{mixmatch_nips19}. Therefore, we utilize a new branch with weak augmentation to obtain $\Bar{x}_i$ (top of Figure~\ref{fig:fig3}), which is augmented by only randomly resized cropping and random horizontal flipping.
To make the differences between samples more transparent, we use a smaller temperature to sharpen the distribution.
Specifically, the distribution obtained by $v_i$ is regarded as the online distribution for gradient back-propagation while the distribution obtained by $\Bar{v}_i$ is used as the target distribution.
Accordingly, for the $i$-th instance sampled from the min-batch, the online distribution $S^{o}(i)$ and target distribution $S^{t}(i)$ can be calculated by $
    S^{o}(i) = \{s_{j}=v_{i} \cdot \Tilde{v}_{j}^{T} / \tau_{ot} | j=1,2,3,...,N\}$
and $
    S^{t}(i) = \{s_{j}=\Bar{v}_{i} \cdot \Tilde{v}_{j}^{T} / \tau_{tt} | j=1,2,3,...,N\}$ respectively.
Note that $\Tilde{v}_{j}$ denotes the feature stored in the memory bank, $\tau_{ot}$ and $\tau_{tt}$ are temperature parameters that control the degree of sharpening of the distribution.

\textbf{Distribution Alignment.}
For the two generated distributions, our goal is to align them with a given objective function. 
Since Kullback–Leibler (KL) divergence~\cite{KL} is often used in statistics to measure the degree of difference between two distributions, we use KL divergence by default as the objective function to align the two distributions $S^{o}(i)$ and $S^{t}(i)$.
In this way, the objective function of global distribution relation is formed by
 \begin{equation}
     \mathcal{L}_{global} = \frac{1}{N} \sum_{i=1}^{N} D_{KL}(S^{t}(i)|| {S}^{o}(i)).
     \label{l-intra}
 \end{equation}
Note that $N$ refers to the size of the training set, and $S^{t}(i)$ does not perform the gradient back-propagation.

%%%%%%%%%%%%%%%%%%%%%%%%%%%%%%%%%%%%%%%%%%%%
\subsection{Local Interpolation Relation}
We utilize image mixture strategy to quantitatively simulate apparently similar images.
Existing data mixture strategy~\cite{mixup_2018_iclr,cutmix_2019_iccv} forces the model to behave linearly when dealing with in-between training examples, that is, the image and target are the corresponding linear interpolation.
We exploit this linearity to model local interpolation relation.
Specifically, we interpolate image pairs and their features with the same ratio, and then pull the extracted features of the interpolated images and the corresponding interpolated features as close as possible in the feature space. The interpolation consistency relation can be well assimilated by transferring the interpolation ratio from pixel space to feature space. The procedure of local interpolation relation can be detailed as three steps: pixel-level interpolation, feature-level interpolation, and interpolation consistency, Figure~\ref{fig:fig3}.

\textbf{Pixel-level Interpolation.}
For each mini-batch, we first sample an interpolation ratio $r$ from the beta distribution as $r \in Beta(\alpha, \alpha)$, where $\alpha$ is a hyper-parameter set to $1.0$ by default.
Then, for two selected instances $x_{i}$ and $x_{j}$ in the mini-batch with size $N_b$, they are interpolated with the ratio $r$ to form the synthetic image $x_{i,j}$. The embedding feature of the interpolated image is defined as
\begin{equation}
    v_{i,j} = f_{\theta}(r \cdot x_{i} \oplus (1-r) \cdot x_{j}),
\end{equation}
where $\oplus$ denotes image interpolation operation. 
Specifically, to randomly select two images for interpolating, the index $i$ is sampled from an ordered set $\mathcal{N}_{order}^{N_b}=\{0, 1, 2, ..., N_{b}-1\}$ and $j$ is sampled from a random-arrangement set $\mathcal{N}_{rand}^{N_b}=randperm(\mathcal{N}_{order}^{N_b})$, where $randperm$() denotes shuffle the order randomly.

\textbf{Feature-level Interpolation.} 
To correspond to the feature of interpolated image $v_{i,j}$ under the simple linearization constraint~\cite{mixup_2018_iclr,cutmix_2019_iccv}, we generate interpolated feature $\Tilde{v}_{i,j}$ according to the ratio $r$, which is regarded as the pseudo ``ground-truth" feature of $v_{i,j}$.
The normalized feature interpolation in the embedding space can be obtained by
\begin{equation}
    \Tilde{v}_{i,j} = \ell_{2}(r \cdot  f_{\theta}(x_{i}) + (1-r) \cdot  f_{\theta}(x_{j})),
\end{equation}
where $\ell_{2}$ denotes normalization.

\textbf{Interpolation Consistency.}
To assimilate interpolation consistency relations, the feature of the interpolated image $v_{i,j}$ and the interpolated feature $\Tilde{v}_{i,j}$ should be attracted to each other, that is transferring interpolation ratio from pixel space to feature space.
It can be achieved using contrastive loss. Accordingly, the loss function of the local interpolation relation is formulated as
\begin{equation}
\label{local_loss}
\begin{split}
    \mathcal{L}_{local} &= \frac{1}{N} \sum_{i\in \mathcal{N}_{order}^{N},j\in \mathcal{N}_{rand}^{N}} -log P(i,j),  \\
    P(i,j) &= \frac{exp(v_{i,j} \cdot \Tilde{v}_{i,j}/\tau)}{\sum_{k\in [0,K)}exp(v_{i,j} \cdot \Tilde{v}_{k}/\tau)},
\end{split}
\end{equation}
where $\Tilde{v}_{k}$ is the feature stored in the memory bank and $\Tilde{v}_{i,j}$ conducts stop-gradient operation.

\subsection{Discussion}
We detail the differences between our ReCo and existing distribution-based and interpolation-based CSL methods in the aspect of exploiting instance relations. 
Besides, ReCo further pursues the complementary nature of these two relations in retaining semantic structure, Table~\ref{tab:table_INsubset} and Figure~\ref{fig:fig5}.

\textbf{Distribution-based Methods.}
Distribution-based methods utilize different ways to calculate the similarity distribution and then align the distributions to explore global-level relations.
In specific, CO2~\cite{ICLR21_CO2} utilizes the features of the two branches of MoCo-v2 to obtain the similarity distribution and then uses distribution alignment as a regularization term. 
ReSSL~\cite{ReSSL_nips21} utilizes weak data augmentation to obtain the target distribution and uses a single distribution alignment loss as the optimization objective.
CLSA~\cite{clsa_arxiv21} utilizes stronger and regular data augmentation to obtain two distributions as online distribution and target distribution, respectively.
Instead, ReCo uses weak augmentation to obtain the target distribution for distribution alignment, which is used to constrain the InfoNCE loss. Moreover, ReCo uses interpolation to explicitly model the local relation that apparently similar inputs are close in feature space, which is not considered in existing distribution-based methods.
More details can be referred in Table~\ref{tab:table_comparison_global}.

\textbf{Interpolation-based Methods.}
ReCo quantitatively models the relation of the interpolated data to original inputs in the feature space.
Un-Mix~\cite{Unmix_AAAI22} and MixCo~\cite{arxiv20_mixco} interpolate in the input space and then weight the loss corresponding to the interpolation ratio.
In contrast, we directly interpolate the features according to the interpolation ratio instead of weighting the loss.
In specific, there are 4 options for interpolation: q and $randperm$(q),  q and $randperm$(k), k and $randperm$(q), and k and $randperm$(k), where $randperm$() denotes randomly shuffle the order of the batch.
Since different views may differ greatly in the early training stages, we claim that the choice of image pairs and feature pairs for interpolation has a large impact on the interpolation consistency relation, which has been completely ignored in previous methods.
More importantly, ReCo not only considers the local interpolation relation, but also further explores the global distribution relation.
More comparisons are shown in Table~\ref{tab:table_comparison_local}.

\setlength{\tabcolsep}{3pt}
%%%%%%%%%%%%%%%%%%%%%%%%%
\begin{table}[!t]

\begin{minipage}{\columnwidth}
\centering
\begin{minipage}[t]{0.46\columnwidth}
\centering
\makeatletter\def\@captype{table}\makeatother\caption{Ablation of temperature parameters in distribution generation.}
\label{tab:table_temperature}

\begin{tabular}{llcc}
\toprule
\midrule
$\tau_{tt}$ & $\tau_{ot}$ &LC Top-1 &LC Top-5  \\
   
\midrule
0.2 &0.2 &70.7 &90.6 \\
0.1 &0.2 &70.6 &90.6 \\
% 0.07 &0.2 &- &- \\
0.1 &0.1 &72.5 &91.1 \\
0.07 &0.1 &73.1 &\textbf{92.5} \\
\textbf{0.04} &\textbf{0.1} &\textbf{73.6} &92.4 \\
% 0.02 &0.1 &- &- \\
0.01 &0.1 &72.1 &91.6 \\
0.2 &0.1 &68.9 &89.4 \\

\bottomrule
\end{tabular}
\end{minipage}
\begin{minipage}[t]{0.46\columnwidth}
\centering
\makeatletter\def\@captype{table}\makeatother\caption{Comparison of different settings for the coefficients in the loss function.}
\label{tab:table_coefficient}

\begin{tabular}{llcc}
\toprule
\midrule
$\lambda_{1}$ & $\lambda_{2}$ &LC Top-1 &LC Top-5  \\
\midrule
0.0 &0.0 &66.2 &88.1 \\
   
% \midrule
% 0.2 &0.0 &71.3 &90.8 \\
0.5 &0.0 &72.8 &91.5 \\
1.0 &0.0 &73.6 &92.4 \\
2.0 &0.0 &70.5 &91.3 \\
% \midrule
% 0.0 &0.5 &71.5 &91.9 \\
% 0.0 &1.0 &74.1 &92.4 \\
% \textbf{0.0} &\textbf{2.0} &\textbf{74.9} &\textbf{92.9} \\
% 0.0 &3.0 &61.8 &86.5 \\
% 1.0 &0.5 &77.5 &93.7 \\
1.0 &1.0 &78.1  &94.4 \\
\textbf{1.0} &\textbf{2.0} &\textbf{78.8} &\textbf{95.0} \\
1.0 &3.0 &78.5 &94.7 \\
% 0.5 &1.0 &78.2 &94.6 \\
% 2.0 &1.0 &77.0 &94.3 \\
\bottomrule
\end{tabular}
\end{minipage}
\end{minipage}
\end{table}

\begin{table}
\setlength{\tabcolsep}{9pt}
\caption[]{Comparison of intra-/ inter-class similarity. ($\times$100)}
\begin{center}

\begin{tabular}{llll}
\toprule
\midrule
Methods &$s_{intra}$ ($\uparrow$) &$s_{inter}$ ($\downarrow$) &$\phi$ ($\uparrow$) \\
\midrule
MoCo-v2~\cite{mocov2_arXiv2020} &35.5 &0.5 &134.9 \\
ReCo &40.1 \textcolor[rgb]{0,0.7,0.3}{(+4.6)} &0.5 &139.4 \textcolor[rgb]{0,0.7,0.3}{(+4.5)} \\
\bottomrule
\end{tabular}

\label{tab:table_intra_inter_similarity}
\end{center}
\end{table}

\section{Experiments}

\subsection{Experimental Settings}

\textbf{Dataset.}
ImageNet-1K~\cite{ImageNet-IJCV15} contains 1,281,167 images with 1000 categories for training and 50,000 images for testing. 
We take ImageNet-100~\cite{Tian_2019_CMC} for fast evaluation, which is a subset of ImageNet and contains 126,689 images with 100 categories for training and 5,000 images for testing.
% All images of the ImageNet dataset are resized to 224 $\times$ 224.
PASCAL VOC~\cite{VOC} contains 20 categories of objects. VOC 2007 contains 5,011 images with 12,608 objects in the $trainval$ set and 4,952 images with 12,032 objects in the $test$ set. 
VOC 2012 contains 11,540 images with 27,450 objects in the $trainval$ set, and labels for the $test$ set have not yet been released.

\textbf{Pre-training Settings.}
ResNet-50~\cite{ResNet} is set as the backbone network by default. 
The size of each view is set to 224 $\times$ 224 for ImageNet pre-training. 
We use the SGD optimizer with the momentum of $0.9$ and weight decay of $0.0005$. 
A cosine learning rate scheduler is employed with a base learning rate of $0.03$, and the batch size is $256$. The temperature $\tau$ of InfoNCE loss is $0.2$. The size of the memory bank is $65536$ and the momentum encoder is updated with a parameter of $0.999$.

\subsection{Ablation Study} 

\textbf{Setup.}
To quickly verify the effectiveness under different parameter settings, we conduct ablation experiments on ImageNet-100~\cite{Tian_2019_CMC} with ResNet-50~\cite{ResNet} architecture and train for 100 epochs. We set the batch size to $128$ with the base learning rate of $0.03$. Other experimental settings are the same as those of ImageNet-1K.

%%%%%%%%%%%%%%%%%%%%%%%%%%
\begin{figure*}[!t]
\centering
% width=1.0\columnwidth
\includegraphics[width=0.98\linewidth]{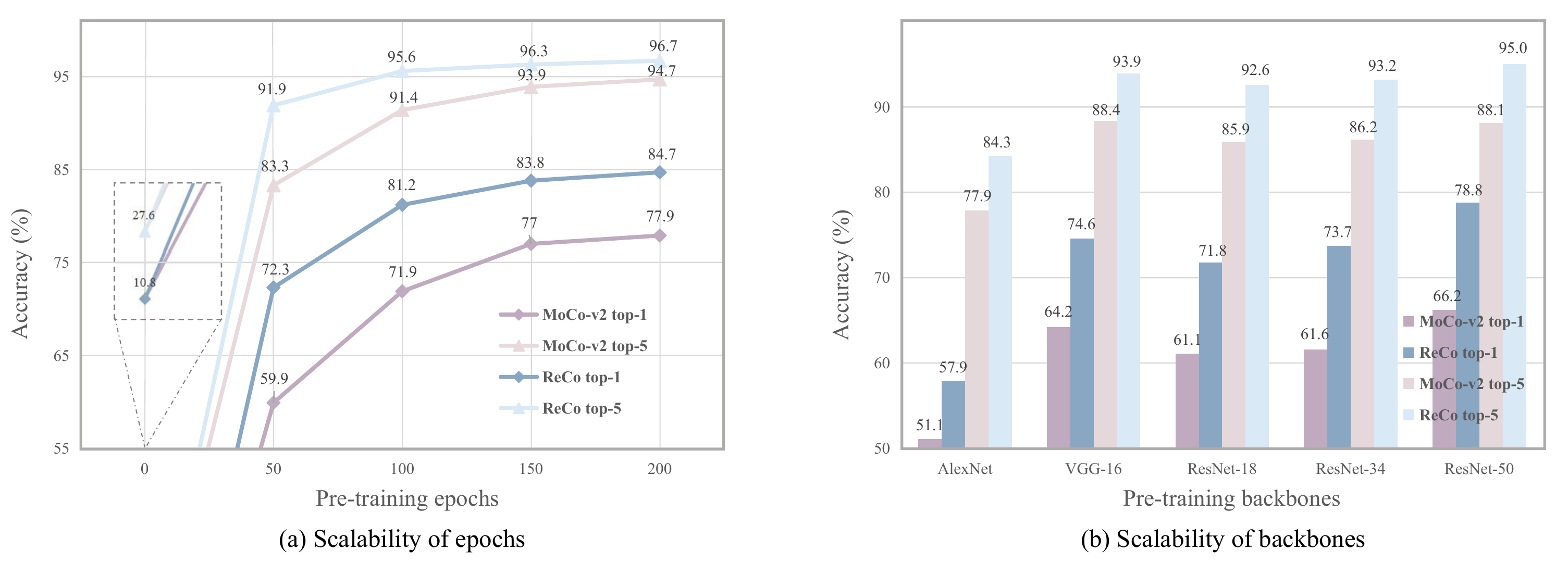}

\caption[]{Scalability of different pre-training epochs and backbones on ImageNet-100. Note that $0$-th epoch in (a) denotes random initialization. 
}
\label{fig:fig4}
\end{figure*}

\textbf{Temperature Parameters.}
For distribution alignment, results of different temperature parameters are shown in Table~\ref{tab:table_temperature}.
In general, performance is better when $\tau_{ot}$ is larger than $\tau_{tt}$. Especially, the performance drops dramatically when $\tau_{tt}$ is larger than $\tau_{ot}$ (the last line).
This is because a smaller $\tau_{tt}$ can sharpen the target distribution and make the difference between various sample pairs more obvious.
We set $\tau_{tt}$=0.04 and $\tau_{ot}$=0.1 by default for the best top-1 accuracy of 73.6\%.

\textbf{Coefficients.}
We study how the global distribution relation term $\mathcal{L}_{global}$ and the local interpolation relation term $\mathcal{L}_{local}$ in Eq. ($\ref{l-all}$) affect the feature representation by using different values of $\lambda_1$ and $\lambda_2$.
When $\lambda_1$=1.0 and $\lambda_2$=2.0, the best top-1 accuracy is 78.8\%.

\setlength{\tabcolsep}{9pt}
\begin{table}
\caption[]{Evaluation on the ImageNet-100 dataset with ResNet-50 by performing linear classification accuracy.}
\begin{center}

\begin{tabular}{lcll}
\toprule
\midrule
Methods &Epochs &LC Top-1 &LC Top-5  \\
   
\midrule
MoCo-v2~\cite{mocov2_arXiv2020}  &100 &66.2 &88.1 \\
BYOL~\cite{BYOL_nips20}  &100 &76.9 &93.8 \\
SimSiam~\cite{SimSiam_CVPR21}  &100 &74.2 &92.8 \\
% MoCo-v2+Un-Mix  &100 &69.5 &90.0 \\
\midrule
MoCo-v2+Local  &100 &74.9 \textcolor[rgb]{0,0.7,0.3}{(+8.7)} &92.9 \textcolor[rgb]{0,0.7,0.3}{(+4.8)} \\
MoCo-v2+Global  &100 &73.6 \textcolor[rgb]{0,0.7,0.3}{(+7.4)} &92.4 \textcolor[rgb]{0,0.7,0.3}{(+4.3)} \\
MoCo-v2+ReCo &100 &78.8 \textcolor[rgb]{0,0.7,0.3}{(+12.6)} &95.0 \textcolor[rgb]{0,0.7,0.3}{(+6.9)}\\
% \midrule
BYOL+Local  &100 &83.0 \textcolor[rgb]{0,0.7,0.3}{(+6.1)} &96.0 \textcolor[rgb]{0,0.7,0.3}{(+2.2)} \\
BYOL+Global &100 &81.2 \textcolor[rgb]{0,0.7,0.3}{(+4.3)} &95.8 \textcolor[rgb]{0,0.7,0.3}{(+2.0)} \\
BYOL+ReCo &100 &83.9 \textcolor[rgb]{0,0.7,0.3}{(+7.0)} &96.7 \textcolor[rgb]{0,0.7,0.3}{(+2.9)} \\

\midrule
    MoCo-v2~\cite{mocov2_arXiv2020}   &200 & 77.9 &94.7\\
    MoCo-v2+ReCo &200 & 84.0 \textcolor[rgb]{0,0.7,0.3}{(+6.1)} &96.3 \textcolor[rgb]{0,0.7,0.3}{(+1.6)}\\
\bottomrule
\end{tabular}

\label{tab:table_INsubset}
\end{center}
% \end{table}
% \setlength{\tabcolsep}{1.4pt}

%%%%%%%%%%%%%%%%%%%
% Table ImageNet-100 detection
% ~\\
\setlength{\tabcolsep}{8pt}
% \begin{table}
\caption[]{Evaluation of object detection on VOC 07+12 with the model pre-trained on ImageNet-100 for 100 epochs. 
$^\dagger$ denotes the modified pre-training settings. 
}
\begin{center}

\begin{tabular}{llll}
\toprule
\midrule
Methods &AP &AP$_{50}$ &AP$_{75}$  \\
   
\midrule
MoCo-v2~\cite{mocov2_arXiv2020}  &48.7 &76.1 &52.4 \\

MoCo-v2+Local  &51.1 \textcolor[rgb]{0,0.7,0.3}{(+2.4)} &77.6 \textcolor[rgb]{0,0.7,0.3}{(+1.5)} &55.3 \textcolor[rgb]{0,0.7,0.3}{(+2.9)} \\
MoCo-v2+Global  &49.0 \textcolor[rgb]{0,0.7,0.3}{(+0.3)} &76.7 \textcolor[rgb]{0,0.7,0.3}{(+0.6)} &52.4 (+0.0) \\

MoCo-v2+ReCo &51.0 \textcolor[rgb]{0,0.7,0.3}{(+2.3)} &78.0 \textcolor[rgb]{0,0.7,0.3}{(+1.9)} &55.3 \textcolor[rgb]{0,0.7,0.3}{(+2.9)} \\
% \midrule
MoCo-v2+ReCo$^\dagger$  &52.0 \textcolor[rgb]{0,0.7,0.3}{(+3.3)} &78.8 \textcolor[rgb]{0,0.7,0.3}{(+2.7)} &56.8 \textcolor[rgb]{0,0.7,0.3}{(+4.4)} \\   %%% lc 78.6
\bottomrule
\end{tabular}

\label{tab:table_INsubsetDetect}
\end{center}
\end{table}

\textbf{Intra-/ Inter-Class Similarity.}
To quantitatively verify the semantic structure of the feature space~\cite{Zhuang_2019_LA,icml20_alignment_uniform,understandeCL_CVPR21}, we define the intra-class similarity as $s_{intra} = \frac{1}{N}\sum_{i}^{N}\sum_{x^+ \in \mathcal{S}_p} \frac{x_i\cdot x^+}{||\mathcal{S}_p||}$, the inter-class similarity as $s_{inter} = \frac{1}{N}\sum_{i}^{N}\sum_{x^- \in \mathcal{S}_n} \frac{x_i\cdot x^-}{{||\mathcal{S}_n||}}$, and the discriminative index as  $\phi = \frac{1}{N}\sum_{i}^{N} \frac{\sum_{x^+ \in \mathcal{S}_p} \frac{x_i\cdot x^+}{||\mathcal{S}_p||} + 1}{\sum_{x^- \in \mathcal{S}_n} \frac{x_i\cdot x^-}{{||\mathcal{S}_n||}} + 1}$, where $N$ is the number of samples,  $\mathcal{S}_p$ is the set of all samples that belong to the same semantic class as $x_i$ based on the ground truth, and $\mathcal{S}_n$ is the set of samples of other different classes. Table~\ref{tab:table_intra_inter_similarity} reports the results on the ImageNet-100 $val$ set. Experimental results show that ReCo presents higher intra-class similarity and discriminative index than MoCo-v2, which demonstrates that a better semantic structure is obtained~\cite{icml19_theory_analysis,Zhuang_2019_LA,understandeCL_CVPR21}.

%%%%%%%%%%%%%%%%%%%%
%%%%%%%%%%%%%%%%%%
% Comparison Global Relation

\begin{table}

\setlength{\tabcolsep}{3pt}
\caption[]{Comparison of the differences between distribution-based methods under training ImageNet-100 for 100 epochs.}
\begin{center}

\begin{tabular}{lccccccl}
\toprule
\midrule
Methods &Dim. &Encod. &Aug. &Sharp. &Contra. &Decoup. &Acc.  \\
\midrule
MoCo-v2~\cite{mocov2_arXiv2020} &128 &o/t &r/r &- &$\checkmark$ &- &66.2 \\
CO2~\cite{ICLR21_CO2}  &128 &o/t &r/r  &- &$\checkmark$ &- &67.5 \\
CLSA~\cite{clsa_arxiv21}  &128 &o/o &s/r  &- &$\checkmark$ &$\checkmark$ &71.1 \\
ReSSL~\cite{ReSSL_nips21} &512 &o/t &r/w  &$\checkmark$ &- &- &72.7 \\
\midrule
ReCo-Global &128 &o/t &r/w  &$\checkmark$ &$\checkmark$ &$\checkmark$ &\textbf{73.6} \\
 
\bottomrule
\end{tabular}

\label{tab:table_comparison_global}
\end{center}
% \end{table}
% \setlength{\tabcolsep}{1.4pt}
% ~ \\
%%%%%%%%%%%%%%%%%%
% Comparison Interpolation

\setlength{\tabcolsep}{3.7pt}
% \begin{table}
\caption[]{Comparison of the differences between interpolation-based methods under training ImageNet-100 for 100 epochs.}
\begin{center}

\begin{tabular}{lcccccccl}
\toprule
\midrule
\multirow{2}{*}{Methods}  &\multicolumn{2}{c}{Image Inter.} &\multicolumn{4}{c}{Feature Inter.} &\multirow{2}{*}{Loss Inter.} &\multirow{2}{*}{Acc.} \\
% \cline{2-7}
&q/q &q/k &q/k &k/q &q/q &k/k \\
\midrule
MoCo-v2~\cite{mocov2_arXiv2020} &- &- &- &- &- &- &- &66.2 \\
MixCo~\cite{arxiv20_mixco} &$\checkmark$ &- &- &- &- &- &$\checkmark$ &69.4 \\
Un-Mix~\cite{Unmix_AAAI22} &$\checkmark$ &- &- &- &- &- &$\checkmark$ &69.5 \\
\midrule
ReCo-Local &$\checkmark$ &- &- &- &$\checkmark$ &- &- &72.4 \\
ReCo-Local &- &$\checkmark$  &$\checkmark$ &-  &- &- &- &\textbf{74.9} \\
ReCo-Local &- &$\checkmark$  &- &$\checkmark$ &- &- &- &71.9 \\
ReCo-Local &- &$\checkmark$  &-  &- &- &$\checkmark$ &- &72.0 \\

\bottomrule
\end{tabular}

\label{tab:table_comparison_local}
\end{center}
\end{table}
% \setlength{\tabcolsep}{1.4pt}

% \textbf{\textcolor{red}{Generalization.}}

%-----------------------------------------------------------------
% Table 1
\setlength{\tabcolsep}{9pt}
\begin{table*}
\caption[]{Comparisons on ImageNet-1K under linear classification (LC) evaluation.
% ``Source" refers to the source of the accuracy in the table.
$^{\dag}$ denotes multi-crop augmentation. $^{\ddag}$ means adding an additional fully connected layer (2048-d, with BN) before the 2-layer MLP. $^*$ denotes our re-implementation.}
\begin{center}

\begin{tabular}{llcccccll}
\toprule
\midrule
Methods &Publisher &Source &Baseline & Architecture &Batch Size &Epochs &LC Top-1 &LC Top-5 \\
% \noalign{\smallskip}
\midrule
Supervised &- &~\cite{mocov2_arXiv2020} &- &R50  &- &90 &76.5 &- \\  

NPID~\cite{Wu_2018_IR} &CVPR18 &~\cite{Wu_2018_IR} &- &R50  &256 &200 &54.0 &-\\
LA~\cite{Zhuang_2019_LA} &ICCV19 &~\cite{Zhuang_2019_LA} &- &R50  &128 &200 &60.2 &-\\
MoCo~\cite{He_2020_Moco} &CVPR20 &~\cite{He_2020_Moco} &- &R50   &256 &200 &60.6 &- \\
% BoWNet~\cite{CVPR2020_BoWNet} &CVPR20 &~\cite{CVPR2020_BoWNet} &- &R50   &256 &200 &60.2 &- \\
% PCL~\cite{PCL_iclr21} &ICLR21 &~\cite{PCL_iclr21} &MoCo &R50  &256 &200 &61.5 &- \\
MoCo-v2~\cite{mocov2_arXiv2020} &arXiv20 &~\cite{mocov2_arXiv2020} &- &R50-MLP  &256 &200 &67.5 &- \\
SimCLR~\cite{Chen_2020_SimCLR} &ICML20 &~\cite{mocov2_arXiv2020} &- &R50-MLP   &8192 &200 &66.6 &- \\
% PIC~\cite{PIC_nips20} &NeurIPS20 &~\cite{PIC_nips20} &- &R50-MLP  &512 &200 &67.3 &87.6 \\
BYOL~\cite{BYOL_nips20} &NeurIPS20 &~\cite{SimSiam_CVPR21} &- &R50-MLP  &4096 &200 &70.6 &- \\
MoCHi~\cite{nips20_MoCHi} &NeurIPS20 &~\cite{nips20_MoCHi} &MoCo-v2 &R50-MLP &256 &200 &68.0 &- \\
MixCo~\cite{arxiv20_mixco} &NeurIPSW20 &~\cite{arxiv20_mixco} &MoCo-v2 &R50-MLP &256 &200 &68.4 &- \\
PCL v2~\cite{PCL_iclr21} &ICLR21 &~\cite{PCL_iclr21} &MoCo-v2&R50-MLP  &256 &200 &67.6 &- \\
CO2~\cite{ICLR21_CO2} &ICLR21 &~\cite{ICLR21_CO2} &MoCo-v2 &R50-MLP  &256 &200 &68.0 &- \\
SimSiam~\cite{SimSiam_CVPR21} &CVPR21 &~\cite{SimSiam_CVPR21} &- &R50-MLP  &256 &200 &70.0 &- \\
JigClu~\cite{jigclu_cvpr21} &CVPR21  &~\cite{jigclu_cvpr21} &- &R50-MLP  &256 &200 &66.4 &- \\
PSL~\cite{cvpr21_psl} &CVPR21 &~\cite{cvpr21_psl} &MoCo-v2 &R50-MLP &256 &200 &68.1 &- \\
ISD~\cite{ICCV21_ISD} &ICCV21 &~\cite{ICCV21_ISD} &BYOL &R50-MLP  &256 &200 &69.8 &- \\
VFT~\cite{ICCV21_VFT} &ICCV21 &~\cite{ICCV21_VFT} &MoCo-v2 &R50-MLP &256 &200 &69.6 &- \\
TKC~\cite{ICCV21_TKC} &ICCV21 &~\cite{ICCV21_TKC} &MoCo-v2 &R50-MLP &256 &200 &69.0 &88.7 \\
ISL~\cite{ICCV21_ISL} &ICCV21 &~\cite{ICCV21_ISL} &MoCo-v2 &R50-MLP &256 &200 &68.6 &- \\
% MSF~\cite{ICCV21_MSF} &ICCV21 &~\cite{ICCV21_MSF} &BYOL &R50-MLP  &256 &200 &72.4 &- \\
ReSSL~\cite{ReSSL_nips21} &NeurIPS21 &~\cite{ReSSL_nips21} &MoCo-v2 &R50-MLP  &256 &200 &69.9 &- \\
CLSA~\cite{clsa_arxiv21} &arXiv21 &~\cite{clsa_arxiv21} &MoCo-v2 &R50-MLP &256 &200 &69.4 &- \\
Un-Mix~\cite{Unmix_AAAI22} &AAAI22 &~\cite{Unmix_AAAI22} &MoCo-v2 &R50-MLP  &256 &200 &68.6 &- \\
HCSC~\cite{HCSC_CVPR22} &CVPR22 &~\cite{HCSC_CVPR22} &MoCo-v2 &R50-MLP  &256 &200 &69.2 &- \\
% \midrule
MoCo-v2$^*$ &arXiv20 &Ours &- &R50-MLP  &256 &200 &67.6 &88.0 \\
% MoCo-v2$^{* \ddag}$ &arXiv20 &Ours &- &R50-MLP  &256 &200 &68.3 &88.7 \\
% BYOL &NeurIPS20 &Ours &- &R50-MLP  &256 &100 &69.7 &89.3 \\
% SimSiam &CVPR21 &Ours &- &R50-MLP  &256 &100 &68.4 &88.3 \\
% MoCo + ReCo &- &Ours &MoCo &R50  &256 &200 &- &- \\
\textbf{ReCo} &- &Ours &MoCo-v2 &R50-MLP  &256 &200 &\textbf{71.3} \textcolor[rgb]{0,0.7,0.3}{(+3.7)} &\textbf{90.5} \textcolor[rgb]{0,0.7,0.3}{(+2.5)} \\
\textbf{ReCo$^{\dag}$} &- &Ours &MoCo-v2 &R50-MLP  &256 &200 &\textbf{73.7} \textcolor[rgb]{0,0.7,0.3}{(+6.1)} &\textbf{91.9} \textcolor[rgb]{0,0.7,0.3}{(+3.9)} \\
% BYOL + ReCo &- &Ours &BYOL &R50-MLP  &256 &200 &- &- \\
\midrule
PIRL~\cite{PIRL_cvpr20} &CVPR20 &~\cite{PIRL_cvpr20} &- &R50  &1024 &800 &63.6 &- \\
SimCLR~\cite{Chen_2020_SimCLR} &ICML20 &~\cite{Chen_2020_SimCLR} &- &R50-MLP  &4096 &1000 &69.3 &89.0 \\
MoCo-v2~\cite{mocov2_arXiv2020} &arXiv20 &~\cite{mocov2_arXiv2020} &-  &R50-MLP  &256 &800 &71.1 &90.1 \\
InvPro~\cite{InvPro_nips20} &NeurIPS20 &~\cite{InvPro_nips20} &NPID &R50-MLP  &128 &800  &71.3 &- \\
PIC~\cite{PIC_nips20} &NeurIPS20 &~\cite{PIC_nips20} &- &R50-MLP  &512 &1600 &70.8 &90.0 \\
SwAV~\cite{SwAv_nips20} &NeurIPS20 &~\cite{SwAv_nips20} &- &R50-MLP  &4096 &400 &70.1 &- \\
SwAV~\cite{SwAv_nips20} &NeurIPS20 &~\cite{SwAv_nips20} &- &R50-MLP  &4096 &800 &71.8 &- \\
InfoMin~\cite{InfoMin_nips20} &NeurIPS20 &~\cite{InfoMin_nips20} &- &R50-MLP  &256 &800 &73.0 &91.1 \\
BYOL~\cite{BYOL_nips20} &NeurIPS20 &~\cite{BYOL_nips20} &- &R50-MLP  &4096 &800 &74.3 &91.6 \\
SwAV$^{\dag}$~\cite{SwAv_nips20} &NeurIPS20 &~\cite{SwAv_nips20} &- &R50-MLP  &4096 &800 &75.3 &- \\
SimSiam~\cite{SimSiam_CVPR21} &CVPR21 &~\cite{SimSiam_CVPR21} &- &R50-MLP  &256 &800 &71.3 &- \\
Barlow Twins~\cite{BarlowTwins_arXiv2021} &ICML21 &~\cite{BarlowTwins_arXiv2021} &- &R50-MLP  &2048 &1000 &73.2 &91.0 \\
NNCLR$^{\dag}$~\cite{ICCV21_NNCLR} &ICCV21 &~\cite{ICCV21_NNCLR} &SimCLR &R50-MLP  &4096 &1000 &75.6 &92.4 \\
VICReg~\cite{VICReg_arXiv2021} &ICLR22 &~\cite{VICReg_arXiv2021} &- &R50-MLP  &2048 &1000 &73.2 &91.1 \\

% \midrule
MoCo-v2$^*$ &arXiv20 &Ours &- &R50-MLP  &256 &800 &70.8 &89.9 \\
\textbf{ReCo} &- &Ours &MoCo-v2 &R50-MLP  &256 &800 &73.7 \textcolor[rgb]{0,0.7,0.3}{(+2.9)} &91.9 \textcolor[rgb]{0,0.7,0.3}{(+2.0)} \\
\textbf{ReCo$^{\dag}$} &- &Ours &MoCo-v2 &R50-MLP  &256 &800 &75.4 \textcolor[rgb]{0,0.7,0.3}{(+4.6)} &92.7 \textcolor[rgb]{0,0.7,0.3}{(+2.8)}  \\
\textbf{ReCo$^{\dag \ddag}$} &- &Ours &MoCo-v2 &R50-MLP  &256 &800 &\textbf{75.9} \textcolor[rgb]{0,0.7,0.3}{(+5.1)}  &\textbf{92.8} \textcolor[rgb]{0,0.7,0.3}{(+2.9)}  \\
\bottomrule

\end{tabular}

\label{tab:table_imagenet}
\end{center}
\end{table*}

\textbf{Scalability.}
The scalability of our model is verified by training with different epochs and backbones. Figure~\ref{fig:fig4} (a) shows the linear classification accuracies of the pre-trained model under different epochs, which shows that higher performance can be obtained with longer training iterations. Moreover, ReCo with 100 epochs can significantly outperform MoCo-v2 with 200 epochs, which demonstrates the pre-training efficiency of ReCo.
Figure~\ref{fig:fig4} (b) further verifies that ReCo can effectively improve the performance of the baseline under various backbones including AlexNet~\cite{alexnet}, VGG-16~\cite{vgg} and ResNet-18/34/50~\cite{ResNet}.

\textbf{Module Efficacy.}
We quantitatively demonstrate the effectiveness of the global distribution relation and local interpolation relation in our ReCo based on MoCo-v2 and BYOL.
In Table~\ref{tab:table_INsubset}, both modules significantly improve the baseline performance, and the combination proves their complementarity for semantic structure retention.
In Table~\ref{tab:table_INsubsetDetect}, the VOC object detection results show that global distribution relation has no obvious advantage in precise location (AP$_{75}$).
We simply set the parameters of the global distribution relation $\tau_{tt}$ and $\tau_{ot}$ to 0.1, AP can be improved by 1.0\%, and AP$_{75}$ by 1.5\%.
This also shows that the pre-trained model performs well on classification do not necessarily perform well on object detection~\cite{analysis_cvpr21}.

\textbf{Distribution-based Methods.}
To demonstrate the difference from existing distribution-based methods, we compare them in detail in Table~\ref{tab:table_comparison_global}.
The differences in related works are reflected in the feature embedding dimension (Dim.), the encoder used to generate the distribution (Encod.), the type of data augmentation (Aug.), whether the distribution is sharpened (Sharp.), whether there is a contrastive learning loss to assist (Contra.), and whether the distribution alignment and contrastive loss are decoupled (Decoup.).
Note that ``o" and ``t" denote the online encoder and target encoder, and ``r", ``w" and ``s" denote regular augmentation, weak augmentation, and strong augmentation respectively.
We reimplement the related methods, and the experimental results show that our global distribution relation (ReCo-Global) achieves the highest performance of 73.6\% on ImageNet-100.

%%%%%%%%%%%%%%%%%%
% Semi supervised

\setlength{\tabcolsep}{4.2pt}
\begin{table}
\caption[]{Comparison of semi-supervised classification. $^\dagger$ denotes official released model. $^*$ denotes our re-implementation.}
\begin{center}

\begin{tabular}{llccccc}
\toprule
\midrule
\multirow{2}{*}{Methods}  &\multirow{2}{*}{Publisher}
&\multirow{2}{*}{Source}
&\multicolumn{2}{c}{1\% label}
&\multicolumn{2}{c}{10\% label} \\
% \cline{2-7}
& & &Top-1 &Top-5 &Top-1 &Top-5 \\
\midrule
NPID~\cite{Wu_2018_IR} &CVPR18 &~\cite{HCSC_CVPR22} &- &39.2 &- &77.4 \\
MoCo-v2~\cite{mocov2_arXiv2020} &arXiv20 &~\cite{HCSC_CVPR22} &36.7 &64.4 &60.7 &83.4 \\
SimCLR~\cite{Chen_2020_SimCLR} &ICML20 &~\cite{HCSC_CVPR22} &46.8 &74.2 &63.6 &86.0 \\
MoCHi~\cite{nips20_MoCHi} &NeurIPS20 &~\cite{HCSC_CVPR22} &38.2 &65.4 &61.1 &83.5 \\
PCL-v2~\cite{PCL_iclr21} &ICLR21 &~\cite{PCL_iclr21} &- &73.9 &- &85.0 \\
CO2~\cite{ICLR21_CO2} &ICLR21 &~\cite{ICLR21_CO2} &- &71.0 &- &85.7 \\
AdCo~\cite{adco_cvpr21} &CVPR21 &~\cite{HCSC_CVPR22} &43.6 &71.6 &61.8 &84.2 \\
HCSC~\cite{HCSC_CVPR22} &CVPR22 &~\cite{HCSC_CVPR22} &48.0 &75.6 &64.3 &86.0 \\
\midrule

HCSC$^\dagger$~\cite{HCSC_CVPR22} &CVPR22 &Ours &48.4  &75.2 &64.0 &86.0 \\
MoCo-v2$^*$~\cite{mocov2_arXiv2020} &arXiv20 &Ours &39.4 &67.8 &61.9 &85.0 \\
\textbf{ReCo} &- &Ours &\textbf{52.8} &\textbf{78.9} &\textbf{66.8} &\textbf{87.9} \\
\bottomrule
\end{tabular}

\label{tab:table_semisup}
\end{center}
% \end{table}
% \setlength{\tabcolsep}{1.4pt}

% ~ \\
\setlength{\tabcolsep}{2.5pt}
% \begin{table}
\caption[]{Comparison of kNN classification performance. $^\dagger$ denotes official released model.}
\begin{center}

\begin{tabular}{lcccccccc}
\toprule
\midrule
\multirow{2}{*}{Methods} 
&\multicolumn{2}{c}{10-NN}
&\multicolumn{2}{c}{20-NN}
&\multicolumn{2}{c}{100-NN}
&\multicolumn{2}{c}{200-NN} \\
% \cline{2-7}
 &Top-1 &Top-5 &Top-1 &Top-5 &Top-1 &Top-5 &Top-1 &Top-5 \\
\midrule
HCSC$^\dagger$~\cite{HCSC_CVPR22} &78.6 &92.5 &78.8 &94.0 &77.6 &95.1 &76.6 &94.9 \\
MoCo-v2~\cite{mocov2_arXiv2020}  &77.8 &92.1 &77.7 &93.5 &76.0 &94.9 &74.9 &94.8 \\
ReCo &\textbf{84.1} &\textbf{94.9} &\textbf{83.6} &\textbf{95.9} &\textbf{82.6} &\textbf{96.8} &\textbf{82.1} &\textbf{97.0} \\

\bottomrule
\end{tabular}

\label{tab:table_knnin100}
\end{center}
% \end{table}

% ~ \\
%%%%%%%%%%%%%%%%%%%
% Table VOC low-shot
\setlength{\tabcolsep}{5.2pt}
% \begin{table}
\caption[]{Evaluation of low-shot classification on VOC2007 using linear SVMs trained on fixed representations. $^\dagger$ denotes our evaluation of the officially released model. $^*$ denotes full re-implementation. Other results are adopted from PCL~\cite{PCL_iclr21}.}
\begin{center}

\begin{tabular}{llccccc}
\toprule
\midrule
Methods &Architecture &k=1 &k=2 &k=4 &k=8 &k=16  \\
   
\midrule
Random &R50 &8.0 &8.2 &8.2 &8.2 &8.5 \\
Supervised &R50 &54.3 &67.8 &73.9 &79.6 &82.3 \\
\midrule
MoCo~\cite{He_2020_Moco} &R50 &31.4 &42.0 &49.5 &60.0 &65.9 \\
PCL~\cite{PCL_iclr21} &R50 &46.9 &56.4 &62.8 &70.2 &74.3 \\
SimCLR~\cite{Chen_2020_SimCLR} &R50-MLP &32.7 &43.1 &52.5 &61.0 &67.1 \\
MoCo-v2~\cite{mocov2_arXiv2020} &R50-MLP &46.3 &58.3 &64.9 &72.5 &76.1 \\
PCL-v2~\cite{PCL_iclr21} &R50-MLP &47.9 &59.6 &66.2 &74.5 &78.3 \\
\midrule
Supervised$^\dagger$ &R50 &54.0 &67.9 &73.8 &79.7 &82.3 \\
ReSSL$^\dagger$~\cite{ReSSL_nips21} &R50-MLP &45.3 &58.1 &66.4 &74.5 &79.3 \\
HCSC$^\dagger$~\cite{HCSC_CVPR22} &R50-MLP &47.9 &59.6 &66.3 &74.4 &78.4 \\
MoCo-v2$^*$~\cite{mocov2_arXiv2020} &R50-MLP &47.1 &58.3 &65.1 &72.4 &76.3 \\
\textbf{ReCo} &R50-MLP &\textbf{54.8} &\textbf{65.8} &\textbf{73.4} &\textbf{79.5} &\textbf{82.6} \\

\bottomrule
\end{tabular}

\label{tab:table_lowshot}
\end{center}
\end{table}
\setlength{\tabcolsep}{1.4pt}

\textbf{Interpolation-based Methods.}
In Table~\ref{tab:table_comparison_local}, our method differs from related works in the implementation of the interpolation method and interpolation ratio correspondence. In particular, previous methods interpolate at the loss level (Loss Inter.), while we interpolate at the feature level (Feature Inter.). In addition, we also compare the impact of different interpolation methods in our local interpolation relation (ReCo-Local).
The q/k interpolation in the corresponding pixel space and feature space obtains the best performance of 74.9\%.

%Table Object Detection
%%%%%%%%%%%%%%%%%%%%%%%%%%%%%%%
%%%%%%%%%%%%%%%%%%%%
\setlength{\tabcolsep}{5pt}
\begin{table}
\caption{Fine-tuning object detection on PASCAL VOC. $^*$ denotes our re-implementation.}
\centering
\begin{tabular}{llccccc}
\toprule
\midrule
Methods  & Publisher &Source & AP & AP$_{50}$ & AP$_{75}$ \\
\midrule
Rand Init  & -  &~\cite{SimSiam_CVPR21} & 33.8 & 60.2 & 33.1 \\ 
Supervised & - &~\cite{mocov2_arXiv2020}   & 53.5 & 81.3 & 58.8 \\
\midrule
MoCo~\cite{He_2020_Moco}       & CVPR20 &~\cite{mocov2_arXiv2020}   & 55.9 & 81.5 & 62.6 \\ %MoCov2 24k
MoCo-v2~\cite{mocov2_arXiv2020}    & arXiv20  &~\cite{mocov2_arXiv2020}   & 57.0 & 82.4 & 63.6 \\ %MoCov2
% SimSiam~\cite{SimSiam_CVPR21}    & CVPR21 &~\cite{SimSiam_CVPR21}    & 56.4 & 82.0 & 62.8 \\ 
CO2~\cite{ICLR21_CO2} &ICLR21 &~\cite{ICLR21_CO2} &57.2 &82.7 &64.1 \\
BarlowTwins~\cite{BarlowTwins_arXiv2021} &ICML21 &~\cite{BarlowTwins_arXiv2021} &56.8 &82.6 &63.4 \\
MaskCo~\cite{MaskCo_iccv21}     & ICCV21 &~\cite{MaskCo_iccv21}    & 56.7 & 82.1 & 63.9 \\ % MaskCo
Un-Mix~\cite{Unmix_AAAI22} &AAAI22 &~\cite{Unmix_AAAI22} & \textbf{57.7} &83.0 &64.3 \\

% VICReg~\cite{VICReg_arXiv2021} &ICLR22 &~\cite{VICReg_arXiv2021} &- &82.4 &- \\
HCSC~\cite{HCSC_CVPR22} &CVPR22 &~\cite{HCSC_CVPR22} &- &82.5 &- \\
ContrastiveCrop~\cite{ContrastiveCrop_CVPR22} &CVPR22 &~\cite{ContrastiveCrop_CVPR22} &57.3 &82.5 &63.8 \\
\midrule
% ReSSL adopts the released model in github
ReSSL$^*$~\cite{ReSSL_nips21} &NeurIPS21 &Ours &55.6 &82.2 &61.6 \\
% MoCo-v2$^*$ &arXiv20 &Ours &57.1 &82.5 &64.3 \\
% ReCo(b=16) &- &Ours &- &- &- \\

% \midrule
MoCo-v2$^*$~\cite{mocov2_arXiv2020} &arXiv20 &Ours &57.1 &82.3 &64.1 \\
\textbf{ReCo} &- &Ours &\textbf{57.7} &\textbf{83.2} &\textbf{64.7} \\

\bottomrule
\end{tabular}
\label{tab:transfer_VOCdet}
% \end{table}

~ \\
~ \\

\setlength{\tabcolsep}{3.5pt}
% \begin{table}
\caption[]{Comparison of object detection and instance segmentation on COCO using Mask R-CNN. $^*$ denotes our reimplementation.}
\centering
\begin{tabular}{lcllllll}
\toprule
\midrule
\multirow{2}{*}{Methods} &\multirow{2}{*}{Source} 
&\multicolumn{3}{c}{Object detection}
&\multicolumn{3}{c}{Instance segmentation} \\

 & & AP$^{bb}$ & AP$_{50}^{bb}$ & AP$_{75}^{bb}$ &AP$^{mk}$ & AP$_{50}^{mk}$ & AP$_{75}^{mk}$ \\
\midrule
Random &~\cite{DetCo_iccv21} &31.0 &49.5 &33.2 &28.5 &46.8 &30.4 \\
Supervised &~\cite{DetCo_iccv21} &38.9 &59.6 &42.7 &35.4 &56.5 &38.1 \\
MoCo-v2~\cite{mocov2_arXiv2020} &~\cite{DetCo_iccv21} &38.9 &59.4 &42.4 &35.5 &56.5 &38.1 \\
DetCo~\cite{DetCo_iccv21} &~\cite{DetCo_iccv21} &39.5 &60.3 &43.1 &35.9 &56.9 &38.6\\
\midrule
MoCo-v2$^*$~\cite{mocov2_arXiv2020} &Ours &39.0 &59.7 &42.7 &35.5 &56.9 &38.0\\
ReCo &Ours &\textbf{39.9} &\textbf{60.8} &\textbf{43.7} &\textbf{36.4} &\textbf{57.6} &\textbf{39.3} \\

\bottomrule
\end{tabular}

\label{tab:transfer_COCOdetseg}
\end{table}

% 逻辑 ImageNet上线性分类和半监督分类，迁移到VOC上少样本分类和目标检测
\subsection{Performance and Comparison}
Comparisons are listed on extensive downstream tasks: linear classification, semi-supervised classification, kNN classification, low-shot classification, object detection, and instance segmentation. 

\subsubsection{Linear Classification}
Convolutional layers initialized by the pre-trained model are frozen while a fully connected linear classifier is initialized from scratch. Its results represent the discriminative ability of the learned representation.

\textbf{Setup.}
We use a LARS optimizer with weight decay of $0.$ and momentum of $0.9$ to train a linear classifier.
A cosine decay schedule is used with an initial learning rate of $0.1*batch/256$ and a batch size of $4096$ for training $90$ epochs. 
In addition, we have tried another setting, which uses the SGD optimizer with a batch size of $256$ for training $100$ epochs.
The learning rate is initialized to $10$ with a decreasing strategy that the rate is scaled down by $0.1$ at $60$-th epoch and $0.01$ at $80$-th epoch.
Both training settings obtain the same performance of 71.3\%, and we use the first one by default.

\textbf{Results.}
Table~\ref{tab:table_imagenet} reports the top-1 and top-5 accuracies of SOTA methods on ImageNet-1K, where our re-implementation of MoCo-v2 achieves 67.6\% top-1 accuracy (0.1\% higher than the official result).
By incorporating instance relations exploration, our ReCo achieves a new SOTA top-1 accuracy of 71.3\%, which improves the baseline MoCo-v2 by 3.7\%.
This demonstrates that ReCo retains data semantic structures via exploring instance relations to enhance the feature discriminative capabilities.
Trained with merely 200 epochs, ReCo even exceeds that of MoCo-v2 with 800 epochs, which proves that ReCo can also improve the pre-training efficiency.
In addition, the performance can be further promoted to 75.9\% when adding the multi-crop data augmentation with 800-epoch training.

%%%%%%%%%%%%%%%%%%%%%%%%%%
\begin{figure}[!t]
\centering
% width=1.0\columnwidth
\includegraphics[width=0.96\columnwidth]{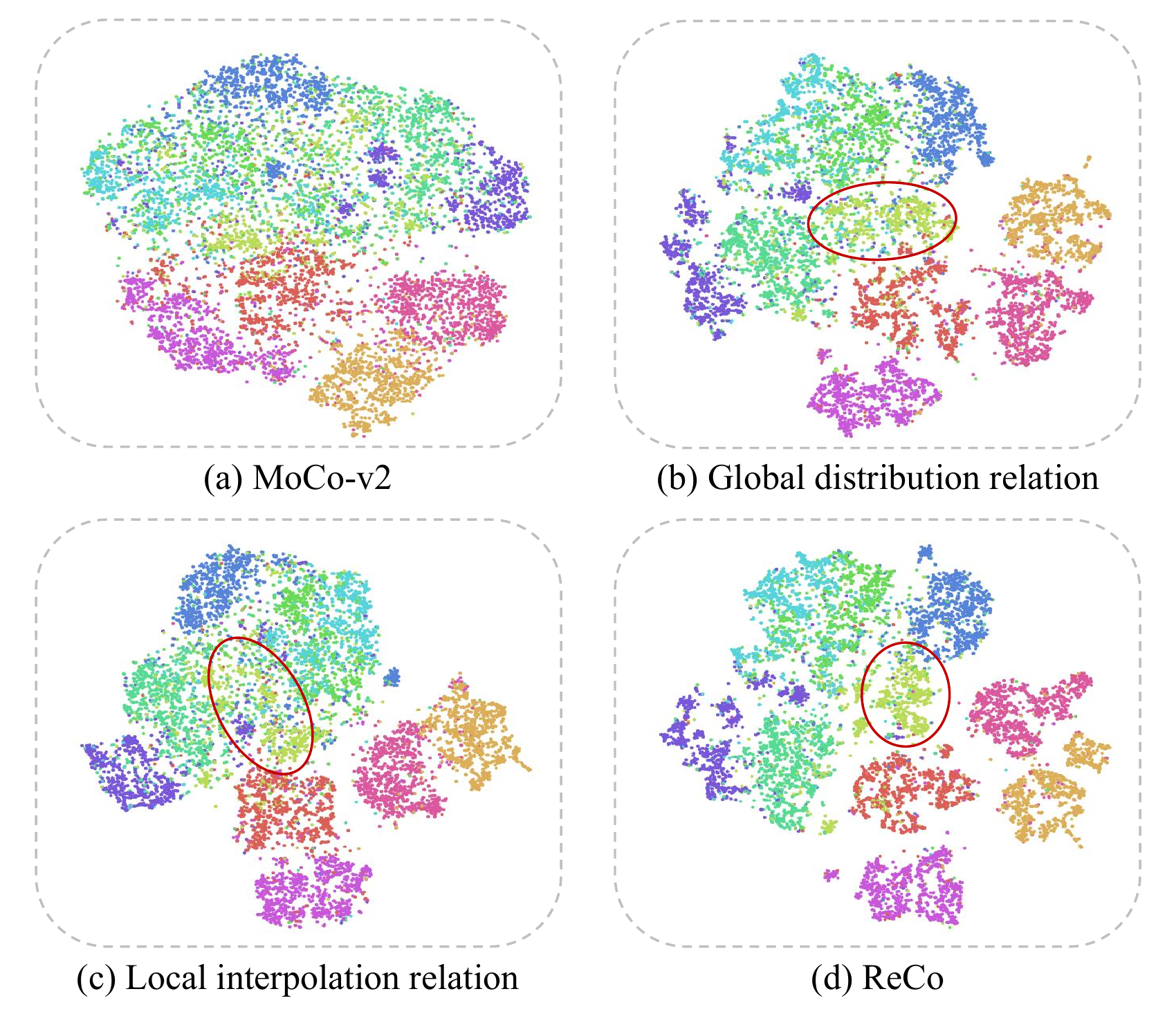}

\caption[]{Visualization of 2-dimensional t-SNE distributions of the embedding space. The area circled in red is more discriminative. (Best viewed in color)

}
\label{fig:fig5}
\end{figure}

%%%%%%%%%%%semi
\subsubsection{Semi-supervised Classification}
Semi-supervised classification first learns from large-scale unlabeled data and then fine-tunes on small labeled data.

\textbf{Setup.}
The backbone and linear layer are fine-tuned on ImageNet with 1\% and 10\% labeled data.
% The SGD optimizer is used to train $20$ epochs with a batch size of $256$. The learning rate is set to $0.01$ for the backbone and $1.0$ for the linear layer.
We fine-tune the pre-trained model for $20$ epochs with the learning rate of $0.01$ for backbone and $1.0$ for linear layer decayed by $0.2$ after $12$ and $16$ epochs. Momentum is set to $0.9$ and weight decay is $1e-4$ for the SGD optimizer.

\textbf{Results.}
Experimental results in Table~\ref{tab:table_semisup} show that ReCo consistently achieves the best performance under different label fractions.
Specifically, ReCo surpasses MoCo-v2 by 13.4\% top-1 accuracy with 1\% labeled data, which demonstrates that the semantic structure learned by exploring instance relations can be more advantageous under insufficient data settings.

\subsubsection{kNN Classification}
We use KNN to evaluate the discriminative capability of the learned features.

\textbf{Setup.}
Under the setting of HCSC~\cite{HCSC_CVPR22}, we evaluate models pre-trained on ImageNet-1K for kNN classification on ImageNet-100, k $\in$ \{10, 20, 100, 200\}.

\textbf{Results.}
Table~\ref{tab:table_knnin100} shows a comparison of top-1 and top-5 accuracy. Experimental results show that ReCo improves baseline with significant margins,  which validates the effectiveness of retaining better semantic structure.

\subsubsection{Transferring to Low-shot Classification}

To verify the discrimination capability of learned features, we train linear SVM using fixed features of $conv5$ under low-shot settings.

\textbf{Setup.}
Following  PCL~\cite{PCL_iclr21}, linear SVM is trained on the VOC~\cite{VOC} 2007 $trainval$ set and tested on the $test$ set. We select $k$ ($k$=1,2,4,8,16) samples from each class for training. Performance is evaluated by mean average precision (mAP).

\textbf{Results.}
As shown in Table~\ref{tab:table_lowshot}, ReCo improves MoCo-v2 by 7.7/7.5/8.3/7.1/6.3 under 1/2/4/8/16 shots. In particular, the performance of ReCo is comparable to the supervised trained model.
These indicate that the features learned by ReCo are sufficiently discriminative and representative.

%%%%%%%%%%%%%%%%%%%%%%%%%%
\begin{figure}[!t]
\centering
% width=1.0\columnwidth
\includegraphics[width=0.95\columnwidth]{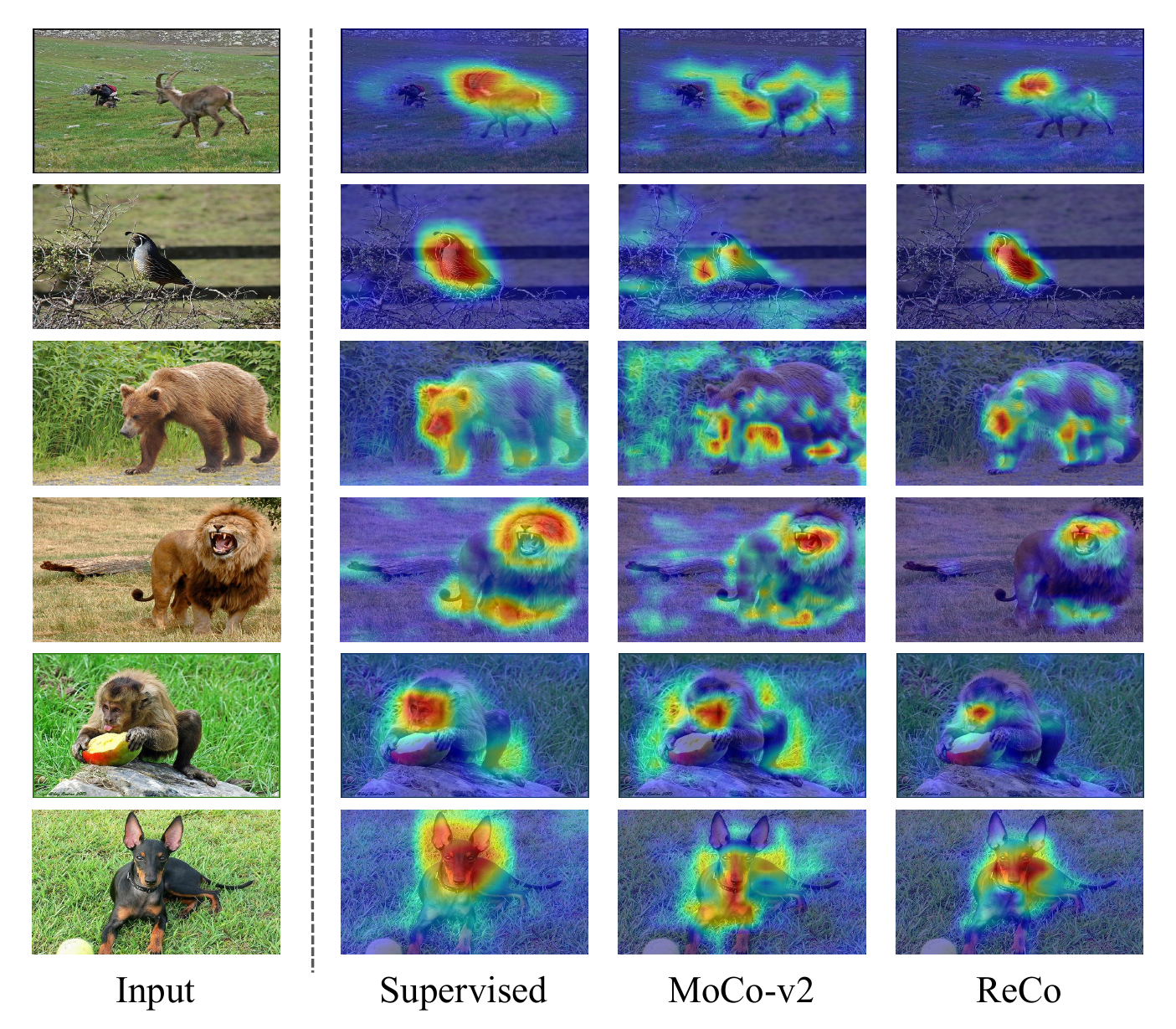}

\caption[]{Activation maps of different pre-trained models using Grad-CAM. Redder colors represent areas that the network pays more attention to.
}
\label{fig:fig6}
\end{figure}

\subsubsection{Transferring to VOC Object Detection.}  
To verify the transferability and generalization capacity of the learned representation, we transfer the trained model to object detection.

\textbf{Setup.}
% Object detection
Following MoCo~\cite{He_2020_Moco},
we fine-tune the Faster R-CNN~\cite{fasterrcnn} detector with ResNet50-C4 architecture on VOC
$trainval$ 07+12 and evaluate the results on $test$ 2007. 
All layers are fine-tuned end-to-end for $48$K iterations with a mini-batch size of $8$ and base learning rate of $0.01$. we set the scale of images to [480, 800] pixels during training and 800 at inference.
The performance of object detection is evaluated by the default VOC metric of AP$_{50}$, as well as COCO-style AP
and AP$_{75}$ as described in MoCo. 

\textbf{Results.}
In Table ~\ref{tab:transfer_VOCdet}, all state-of-the-art CSL methods outperform supervised pre-training on the object detection task, which demonstrates the advantage of CSL for transfer learning.
With $\tau_{tt}$/$\tau_{ot}$=0.1/0.1, ReCo presents 0.6/0.9/0.6 gains over MoCo-v2 under AP/AP$_{50}$/AP$_{75}$.
These results demonstrate that exploring instance relations improves the transferability and generalization of the model.

\subsubsection{Transferring to COCO Object Detection And Instance Segmentation.}
We also evaluate the learned model on large-scale COCO dataset~\cite{COCO}.

\textbf{Setup.}
We fine-tune the Mask R-CNN~\cite{maskrcnn} with ResNet50-FPN architecture on COCO 2017 $train$ set and results are evaluated on $val$ set. We set the batch size to 16 and the learning rate to 0.02 to train for 90K iterations. 

\textbf{Results.}
As shown in Table~\ref{tab:transfer_COCOdetseg}, ReCo yields 1.1\% AP$_{50}^{bb}$ and 1.3\% AP$_{75}^{mk}$ improvements over MoCo-v2 for object detection and instance
segmentation, respectively. These results validate the transferability of the learned model on a variety of tasks.

\subsection{Visualization}

\textbf{Embedding Space.}
We simply use the CIFAR-10~\cite{CIFAR} val set with 10 categories for feature space visualization.
The t-SNE~\cite{tsne2009} technique is utilized to map the feature space onto a 2D plane. 
Figure~\ref{fig:fig5} shows intra- and inter-class variation, which reflects the semantic structure of the data.
With InfoNCE loss, MoCo-v2 can learn semantic structure to a certain extent, but the interstice between different categories is not clear enough.
By considering global distribution relations or local interpolation relations separately, the degree of discrimination of different categories is more obvious than MoCo-v2.
In particular, ReCo can obtain a feature space with better semantic structure by combining these two.

\textbf{Activation Map.}
We use Grad-CAM~\cite{iccv17_gradcam} to visualize activation map.
As shown in Figure~\ref{fig:fig6}, the supervised pre-trained model focuses on the entire object or discriminative regions of the object, while the model learned by MoCo-v2 is more distracted and even focuses on non-foreground object regions.
This is because instance discrimination approaches aim at learning sample-specific features while supervised training exploits semantic label information to learn class-specific discriminative features.
Compared with MoCo-v2, the model learned by ReCo pays more attention to foreground objects, which is more similar to the supervised training model. This demonstrates the advantage of ReCo in retaining semantic structure.

\section{Conclusion}
In this paper, we explicitly exploit semantic relations among instances for relation-aware contrastive self-supervised learning (ReCo).
Unlike previous instance discrimination-based CSL methods that only contrast samples with pre-defined hard binary error-prone assignments, ReCo simultaneously explores the soft relation in instance similarity distributions at the global level and interpolation consistency at the local level.
With a better semantic structure, the learned feature space appears to be locally aggregated yet globally uniform.
%In the future, we will focus more on relating different instances of the same semantic category based on neighborhood discovery or clustering for unsupervised learning.
We expect that ReCo can provide fresh insights into the CSL community, \textit{e.g.}, introducing neighborhood discovery or clustering techniques for better semantic-aware instance relation exploration.

% \section*{Acknowledgement}
% This work is supported by the Beijing Municipal Science \& Technology Commission (Z191100007119002), the Key Research Program of Frontier Sciences, CAS, Grant NO ZDBS-LY-7024.

\ifCLASSOPTIONcaptionsoff
  \newpage
\fi

% \end{thebibliography}
\bibliographystyle{IEEEtran}
\bibliography{main}

\end{document}